\documentclass{article}

\usepackage{microtype}
\usepackage{graphicx}
\usepackage{subcaption}
\usepackage{booktabs} % for professional tables

\usepackage{hyperref}

\usepackage[accepted]{icml2026}

\usepackage{amsmath}
\usepackage{amssymb}
\usepackage{mathtools}
\usepackage{amsthm}

\usepackage{multirow}

\usepackage[capitalize,noabbrev]{cleveref}

\usepackage[utf8]{inputenc}
\usepackage{xcolor}
\usepackage{fontawesome5} % 用于显示图标 (如 \faUser, \faBrain)
\usepackage[most]{tcolorbox} % 用于漂亮的文本框

\definecolor{safety_col}{HTML}{D1495B} % Red-ish
\definecolor{moral_col}{HTML}{4C956C}  % Green-ish
\definecolor{bg_safety}{HTML}{FADBD8}  % 浅红色背景
\definecolor{bg_moral}{HTML}{D4EFDF}   % 浅绿色背景

\newtcolorbox{casebox}[3]{
  enhanced,
  colback=white,       % 内容背景白
  colframe=#2,         % 边框颜色
  colbacktitle=#2,     % 标题背景颜色
  coltitle=white,      % 标题文字颜色
  fonttitle=\bfseries\small,
  title={#1},
  arc=2mm,             % 圆角
  boxrule=0.8pt,       % 边框粗细
  left=1mm, right=1mm, top=1mm, bottom=1mm, % 内部边距
  titlerule=0mm,
}

\newcommand{\casetag}[2]{\textbf{\textcolor{#1}{#2}}}

\theoremstyle{plain}
\newtheorem{theorem}{Theorem}[section]
\newtheorem{proposition}[theorem]{Proposition}
\newtheorem{lemma}[theorem]{Lemma}
\newtheorem{corollary}[theorem]{Corollary}
\theoremstyle{definition}
\newtheorem{definition}[theorem]{Definition}
\newtheorem{assumption}[theorem]{Assumption}
\theoremstyle{remark}
\newtheorem{remark}[theorem]{Remark}

\usepackage[textsize=tiny]{todonotes}

\icmltitlerunning{Towards Context-Invariant Safety Alignment for Large Language Models}

\begin{document}

\twocolumn[
  \icmltitle{Towards Context-Invariant Safety Alignment for Large Language Models}



  \icmlsetsymbol{equal}{*}
  \icmlsetsymbol{intern}{$\dagger$}

  \begin{icmlauthorlist}
    \icmlauthor{Yixu Wang}{fd,sh}
    \icmlauthor{Yang Yao}{sh}
    \icmlauthor{Xin Wang}{fd,sh}
    \icmlauthor{Yifeng Gao}{fd}
    \icmlauthor{Yan Teng}{sh}
    \icmlauthor{Xingjun Ma}{fd,sh}
    \icmlauthor{Yingchun Wang}{sh}
  \end{icmlauthorlist}

  \icmlaffiliation{fd}{Fudan University, Shanghai, China}
  \icmlaffiliation{sh}{Shanghai Artificial Intelligence Laboratory, Shanghai, China}

  \icmlcorrespondingauthor{Yan Teng}{tengyan@pjlab.org.cn}
  \icmlcorrespondingauthor{Xingjun Ma}{xingjunma@fudan.edu.cn}

  \icmlkeywords{Machine Learning, ICML}

  \vskip 0.3in
]



\printAffiliationsAndNotice{}  % no special notice (required even if empty)

\begin{abstract}
Preference-based post-training aligns LLMs with human intent, yet safety behavior often remains brittle. 
A model may refuse a harmful request in a standard prompt but comply when the same intent is wrapped in adversarial wording. 
We suggest that robust safety requires \emph{context-invariant} alignment, where behavior depends on the underlying intent rather than surface form. 
Enforcing invariance is difficult in alignment because not all training signals are equally trustworthy; for some prompt variants we can obtain verifiable feedback (e.g., multiple-choice), while for open-ended variants we typically rely on noisy, gameable reward proxies (e.g., learned judges). 
As a result, standard symmetric invariance regularizers can reduce cross-context discrepancies by lowering performance on reliable variants instead of improving open-ended robustness.
To address this, we introduce \textbf{Anchor Invariance Regularization (AIR)}, which treats verifiable prompts as anchors and uses a \textbf{stop-gradient target} to regularize only the open-ended variants toward the anchor performance. 
AIR is implemented as a plug-in auxiliary loss and combined with group-based preference optimization (e.g., GRPO) via heterogeneous prompt grouping. Across Safety, Moral Reasoning, and Math, AIR improves context invariance, boosting in-distribution group accuracy by 12.71\% and out-of-distribution consistency by 33.49\%, making safety constraints robust to adversarial framings.
\end{abstract}

\section{Introduction}
Preference-based post-training has established itself as the standard paradigm for aligning large language models (LLMs) with human intent~\cite{christiano2017deep,ouyang2022training,bai2022training}. 
However, despite the remarkable success of reinforcement learning from human feedback (RLHF)~\cite{bai2022training,stiennon2020learning,nakano2021webgpt}, safety alignment remains compromised by fragility~\cite{huang2024flames,gu2024mllmguard,yao2025mousetrap}. 
A model may refuse a harmful request in a standard prompt but fail when the same intent is wrapped in adversarial wording. 
This brittleness stems from the optimization landscape itself. 
Because training typically maximizes proxy rewards on isolated prompts, models are incentivized to exploit spurious surface features correlated with high scores, often leading to reward hacking~\cite{skalse2022defining} and alignment faking~\cite{wang2024fake,greenblatt2024alignment}, rather than internalizing the underlying safety constraints. 
Consequently, a model might correctly reject a harmful query in its canonical form, yet comply with the \textit{exact same intent} under jailbreak-style framing~\cite{wei2023jailbroken,zou2023universal,anil2024many,song2025jailbound,ma2026safety}. 
This phenomenon highlights a critical gap: \emph{robust safety behavior should be grounded in the underlying intent, rather than the surface context.}

To move beyond such superficiality, we argue that robust safety alignment requires \emph{Context-Invariance}. 
For the \emph{same underlying intent}, safety behavior should remain stable across different surface contexts (e.g., wording, format, or jailbreak wrappers). 
While invariance constraints are a staple of domain generalization~\cite{arjovsky2019invariant,krueger2021out}, applying them to safety alignment faces a key obstacle, i.e., \emph{supervision is asymmetric across contexts}. 
Open-ended generation rarely has ground truth and relies on noisy, gameable reward proxies~\cite{gao2023scaling,moskovitz2023confronting}, whereas restricted formats (e.g., multiple-choice or rule-checkable constraints) provide verifiable supervision. 
This asymmetry undermines traditional \emph{symmetric} invariance penalties that treat all contexts equally; shrinking cross-context gaps can be achieved by degrading the reliable anchors to match noisier variants, rather than improving open-ended robustness. 
Therefore, enforcing context-invariance in safety alignment calls for an explicitly \emph{asymmetric} mechanism that preserves anchor competence while regularizing open-ended behavior.

In this work, we introduce \textbf{Anchor Invariance Regularization (AIR)}, a reinforcement learning approach that enforces robustness under asymmetric supervision. 
AIR breaks the symmetry of standard invariance objectives by designating verifiable contexts as \textbf{anchors}. 
Rather than forcing all contexts to collapse to a shared average, AIR regularizes open-ended generation to match the anchor reference via a \textbf{stop-gradient target}, ensuring the reliable anchor remains fixed. 
This anchored constraint acts as a bidirectional rectifier. 
It suppresses updates driven by spuriously high open-ended rewards, while encouraging improvements when open-ended behavior falls short of the verifiable standard. 
By leveraging a small fraction of reliable supervision, AIR constrains optimization on noisy, unverifiable prompts without requiring expensive human annotation.

We practically instantiate \textbf{AIR} by integrating it with group-based
preference optimization (e.g., GRPO~\cite{shao2024deepseekmath}). 
To enable efficient invariance estimation within this group-based approach, we introduce a \textbf{heterogeneous prompt grouping} mechanism that constructs training batches containing both verifiable anchors and their corresponding open-ended variants. 
We further formulate AIR as an \textbf{optimizer-agnostic auxiliary loss}, allowing the gradient estimator to directly rectify policy updates based on the anchor-open discrepancy. 
We evaluate this method across a diverse set of domains (i.e., Safety, Moral Reasoning, and Mathematics) to demonstrate its efficacy beyond standard safety protocols. 
These experiments confirm that AIR effectively mitigates alignment faking across a spectrum ranging from objective reasoning to subjective value alignment. 
Empirically, AIR substantially outperforms strong baselines, improving in-distribution group accuracy by 12.71\% and achieving a 33.49\% gain in out-of-distribution consistency.
In summary, our main contributions are as follows:
\begin{itemize}
    \item  We identify that symmetric invariance regularization fails in safety alignment. To resolve this, we propose \textbf{Anchor Invariance Regularization (AIR)}, an asymmetric objective that anchors open-ended generation to verifiable constraints.
    \item We develop a practical reinforcement learning approach that integrates AIR into Group Relative Policy Optimization (GRPO). By leveraging \emph{heterogeneous prompt grouping} and a tailored gradient estimator, our method effectively suppress reward hacking.
    \item We evaluate AIR across Safety, Moral Reasoning, and Math, demonstrating that enforcing context-invariance significantly improves robustness. Notably, AIR achieves a 33.49\% gain in out-of-distribution consistency over state-of-the-art baselines.
\end{itemize}

\section{Related Work}

\textbf{Preference-based Post-training.} \quad
Aligning LLMs with human intent is commonly framed as preference-based post-training~\cite{christiano2017deep,ouyang2022training,wang2026safevid}.
Foundational approaches learn a reward model from pairwise human preferences and optimize the policy via reinforcement learning~\cite{schulman2017proximal}.
To mitigate the complexity and instability inherent in actor-critic frameworks, a growing body of work explores direct or implicit preference optimization. This includes DPO~\cite{rafailov2023direct} and its reference-free or single-stage successors such as ORPO~\cite{hong2024orpo}, KTO~\cite{ethayarajh2024kto}, and SimPO~\cite{meng2024simpo}, which derive policy updates directly from preference data.
In parallel, group-based optimizers (e.g., GRPO~\cite{shao2024deepseekmath}) emerge as a lightweight alternative to critic-based baselines, estimating relative advantages within cohorts of sampled completions.
However, most preference-optimization methods still rely on the assumption that the scalar reward signal is a sufficiently reliable proxy for the latent objective~\cite{gao2023scaling,skalse2022defining,wang2025safeevalagent}.

\textbf{Invariant Risk Minimization.} \quad
Invariant Risk Minimization (IRM)~\cite{arjovsky2019invariant} seeks predictors that remain optimal across multiple environments, discouraging reliance on spurious correlations that only hold in specific settings.
In our alignment setting, we use \emph{contexts} to denote different surface realizations of the same latent intent, which correspond to \emph{environments} in the IRM literature.
Practical variants include gradient-penalty relaxations such as IRMv1 and risk-based surrogates such as Risk Extrapolation (REx) and V-REx~\cite{krueger2021out}, which penalize dispersion of environment risks as a simple alternative to explicit gradient constraints.
While IRM-style objectives have been explored under symmetric~\cite{choe2020empirical,wang2022generalizing} and reliable supervision~\cite{sun2023evaluating,zheng2023invariant,zheng2024improving}, preference-based post-training faces asymmetric reward reliability across contexts.
We therefore adopt a V-REx style objective but break its symmetry, using verifiable variants as anchors to regularize open-ended generation toward high-reliability constraints.

\textbf{Learning from Unverifiable Rewards.} \quad
A central challenge for aligning models is that many objectives are hard to verify, so practice relies on proxy reward signals that can be gamed through reward hacking.
A substantial body of work studies this phenomenon, including scaling behavior and empirical analyses of reward overoptimization, as well as methods that explicitly constrain optimization to mitigate reward hacking~\cite{gao2023scaling,rafailov2024scaling,moskovitz2023confronting,wolf2025reward}.
To improve reward reliability, prior work explores rule-based rewards~\cite{mu2024rule} and LLM-as-a-judge pipelines for evaluating and training on open-ended tasks~\cite{zheng2023judging,chiang2024chatbot,liu2023g,gu2024surveyllmasajudge}.
Reward model benchmarks further suggest that reward quality is non-trivial to assess and may correlate weakly with downstream improvements~\citep{lambert2025rewardbench,mac2025rethinking}.
Finally, weak-to-strong generalization studies motivate using simpler supervision to elicit stronger capabilities, while highlighting failure modes when proxy supervision is miscalibrated~\cite{burns2023eliciting}.
In this work, we address this supervision asymmetry by utilizing verifiable tasks as reliable anchors to constrain the optimization of open-ended generation, thereby preventing overfitting to noisy proxies.
\section{Proposed Method}
In this section, we introduce Anchor Invariance Regularization (AIR) for context-invariant safety alignment.
We first formalize alignment as an invariance problem, then analyze why symmetric V-REx-style penalties fail, and present AIR and its practical implementation as an auxiliary loss.

\subsection{Problem Formulation}
Let $z \sim \mathcal{Z}$ denote a \textbf{latent intent} (e.g., a specific safety constraint), and let $\mathcal{C}$ be a set of \textbf{prompt contexts} (environments in IRM) representing distinct surface realizations of that intent (e.g., verifiable multiple-choice vs.\ open-ended scenarios).
A rendering function $g(z, c)$ maps an intent and context to an observable prompt $s = g(z, c)$.
The policy $\pi_\theta(y \mid s)$ generates a response $y$, evaluated by the expected risk (negative reward) under context $c$:
\begin{equation}
R_c(\theta) = \mathbb{E}_{z \sim \mathcal{Z}} \mathbb{E}_{y \sim \pi_\theta(\cdot \mid g(z, c))} \big[ -r(g(z, c), y, c) \big].
\label{eq:risk_def}
\end{equation}
Standard optimization minimizes the aggregate risk $\sum_{c \in \mathcal{C}} R_c(\theta)$. 
However, this allows models to exploit contexts whose rewards are noisy or gameable, while failing in rigorous, verifiable contexts. Instead, we seek a policy whose optimality is invariant across all $c \in \mathcal{C}$.

\subsection{Naive V-REx Invariance Regularization}
\label{sec:naive-irm}

A natural starting point for enforcing consistency is V-REx~\cite{krueger2021out}, which minimizes the average risk while explicitly penalizing the dispersion of risks across contexts. We formulate the objective as:
\begin{equation}
\mathcal{L}_{\text{naive}}(\theta) = \underbrace{\frac{1}{|\mathcal{C}|}\sum_{c\in \mathcal{C}} R_c(\theta)}_{\text{Average Risk}} + \lambda \cdot \underbrace{\mathrm{Var}_{c\in\mathcal{C}}\big[R_c(\theta)\big]}_{\Omega_{\text{var}}(\theta)},
\label{eq:vrex}
\end{equation}
where $\Omega_{\text{var}}(\theta)$ represents the invariance regularization term. However, we identify a critical failure mode in this naive implementation when applied to asymmetric alignment tasks. To expose this issue, we examine the gradient of the variance penalty. For a simplified case with two contexts ($c_1, c_2$), the gradient is proportional to their gap multiplied by the difference of their gradients:
\begin{equation}
\nabla_\theta \Omega_{\text{var}} \propto (R_1 - R_2) \cdot \big(\nabla_\theta R_1 - \nabla_\theta R_2\big).
\end{equation}
The fundamental flaw lies in the \textbf{symmetry} of this update. To minimize the variance, the optimizer is mathematically permitted to move in \emph{either} direction: it can improve the worse context, or it can degrade the better context.

\begin{figure}
    \centering
    \includegraphics[width=\columnwidth]{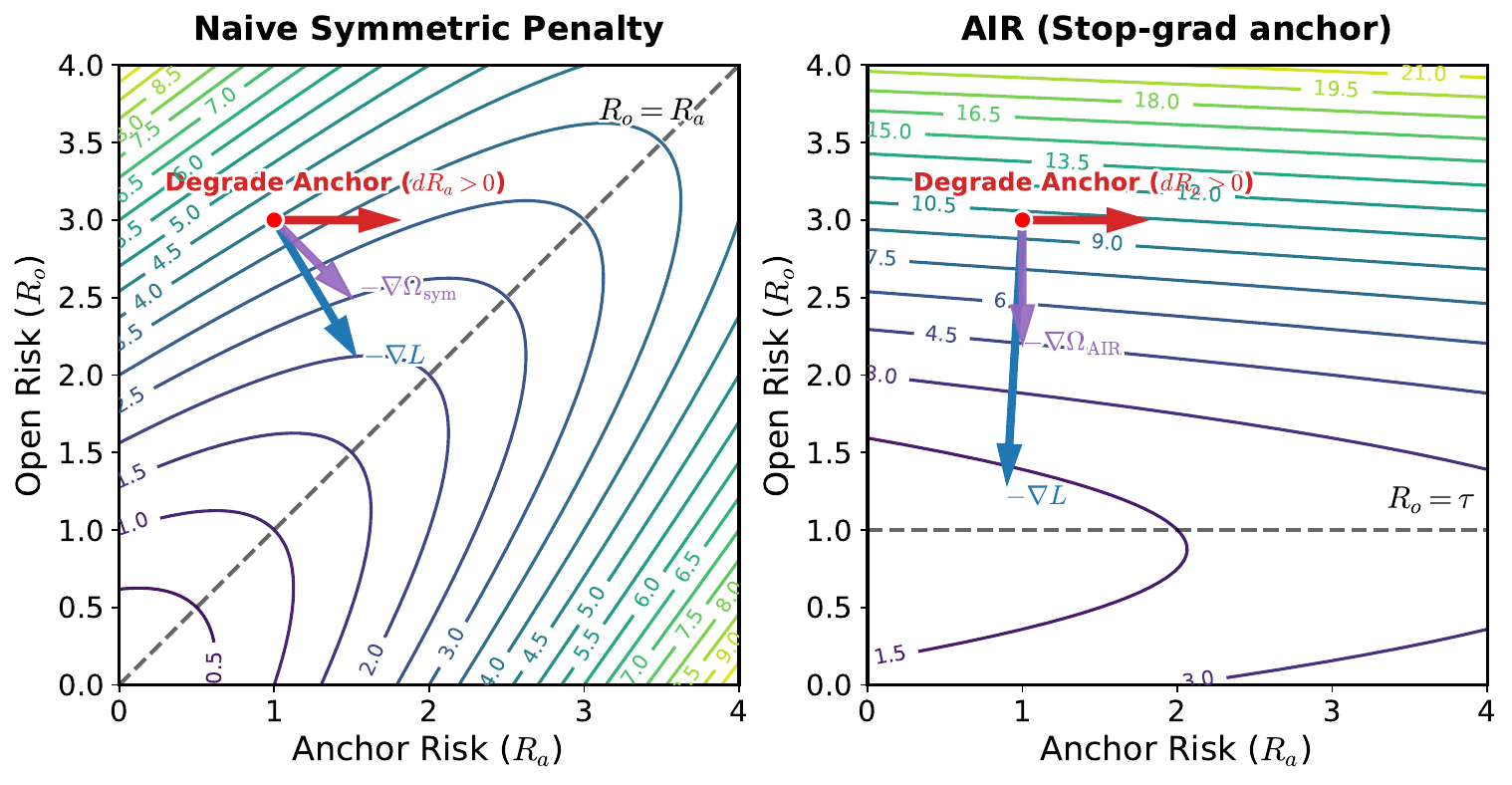}
    \caption{Risk-space geometry of symmetric vs. anchored regularization. While the naive symmetric penalty (left) minimizes variance by degrading the reliable anchor to match the poorer context, AIR (right) breaks this symmetry via a stop-gradient operator. This forces open-ended tasks to align with verifiable competence without compromising the anchor's performance.}
    \vspace{-10pt}
    \label{fig:air}
\end{figure}

Consider the alignment setting where $c_1$ is an \textbf{easy} (verifiable) context and $c_2$ is a \textbf{hard} (noisy) context. As training proceeds, the model naturally performs better on the easy task ($R_1 \ll R_2$). The variance penalty then generates a gradient component acting on $c_1$:
\begin{equation}
\text{Force on Easy} \propto - (R_2 - R_1) \nabla_\theta R_1.
\end{equation}
Under gradient descent, the parameter update contributed by this term is
$\Delta\theta \propto -\nabla_\theta \Omega_{\text{var}}
= - (R_1-R_2)(\nabla_\theta R_1-\nabla_\theta R_2)$.
When $R_1 \ll R_2$, we have $(R_1-R_2)<0$, hence the update contains a component
$\Delta\theta_{\text{easy}} \propto -(R_2-R_1)\nabla_\theta R_1$,
which moves parameters in the direction that \emph{increases} $R_1$ (i.e., degrades the easy context), thereby reducing the gap by collapsing performance. 
This geometric pathology is visualized in Figure~\ref{fig:air} (Left), where the vector field explicitly shows the gradient pushing the anchor risk ($R_a$, corresponding to $R_1$) upwards to minimize the symmetric gap. We provide a formal proof of this failure mode in Appendix~\ref{app:degenerate_existence}.

\subsection{Anchor Invariance Regularization (AIR)}
\label{sec:air}

To make invariance useful for preference-based post-training, we must break this symmetry by \emph{designating a trusted reference context} and only regularizing the others toward it.

\textbf{Anchors as reliable context.} \quad
We assume that for each latent instance $z$ there exists at least one context whose supervision is comparatively reliable (nearly unambiguous and automatically verifiable), e.g., multiple-choice or rule-checkable formats.
We call such a context the \textbf{anchor} and denote it by $c_{\mathrm{acr}}$.
All other context $c\in\mathcal{C}\setminus\{c_{\mathrm{acr}}\}$ are \emph{open}  variants that may rely on noisy rubric-based or model-based evaluation.
Intuitively, we want the policy to preserve competence on the anchor while preventing open context from drifting into prompt-specific shortcuts.

\textbf{The AIR objective.} \quad
AIR replaces the symmetric variance penalty with an anchor-referenced penalty applied exclusively to non-anchor contexts:
\begin{equation}
\quad \Omega_{\text{AIR}}(\theta)=
\sum_{c\in \mathcal{C}\setminus\{c_{\mathrm{acr}}\}}
\big(R_c(\theta)-\tau_{\mathrm{acr}}(\theta)\big)^2,
\end{equation}
where the anchor reference level is defined as:
\begin{equation}
\tau_{\mathrm{acr}}(\theta) = \operatorname{sg}\!\big[\,R_{c_{\mathrm{acr}}}(\theta)\,\big].
\label{eq:tau}
\end{equation}
The stop-gradient operator $\operatorname{sg}[\cdot]$ is essential. It ensures that the gradient of the penalty is zero with respect to the anchor's own performance. Mathematically, $\nabla_\theta \tau_{\mathrm{acr}}(\theta)=0$ within the AIR term, meaning the regularizer cannot reduce the gap by degrading anchor performance. It can only do so by updating the \emph{open} contexts to match the verifiable anchor standard. 
As illustrated in Figure~\ref{fig:air} (Right), this effectively rectifies the optimization geometry. 
The gradient vectors are constrained to reduce the open risk $R_o$ towards the anchor level, without exerting any force that would increase the anchor risk $R_a$.
We formally show that this structural change eliminates the anchor-degrading gradient components in Appendix~\ref{app:air_no_anchor_grad}.

\subsection{Implementation as a Policy-Gradient Auxiliary Loss}

We implement AIR as an optimizer-agnostic policy-gradient auxiliary loss, computed over grouped rollout batches, and add it to the base policy-optimization objective.
Recall the AIR objective $\sum (R_c(\theta) - \tau_{\text{acr}})^2$. Since the risk $R_c(\theta)$ is an expectation over the policy's generations, we cannot directly backpropagate through the risk value. Instead, we apply the log-derivative trick to estimate the gradient of the squared penalty. For a non-anchor context $c$, the gradient of the AIR term is:
\begin{equation}
\begin{split}
\nabla_\theta \Omega_{\text{AIR}, c} 
&= \nabla_\theta \big( R_c(\theta) - \tau_{\text{acr}}(\theta) \big)^2 \\
&= 2 \big( R_c(\theta) - \tau_{\text{acr}}(\theta) \big) \cdot \nabla_\theta R_c(\theta).
\label{eq:air_grad_deriv}
\end{split}
\end{equation}
Substituting the definition of risk gradient $\nabla_\theta R_c(\theta) = - \mathbb{E}_{y \sim \pi} [r(s,y) \nabla_\theta \log \pi_\theta(y|s)]$, Eq.~\eqref{eq:air_grad_deriv} becomes:
\begin{equation}
\begin{split}
& \nabla_\theta \Omega_{\text{AIR}, c} =\\ & -\mathbb{E}_{y \sim \pi} \Big[ \underbrace{2\big(R_c(\theta)-\tau_{\mathrm{acr}}(\theta)\big) \cdot r(s,y)}_{\text{Invariance Coefficient}} \cdot \nabla_\theta \log \pi_\theta(y|s) \Big].
\end{split}
\end{equation}
This derivation reveals that minimizing the AIR penalty is equivalent to maximizing a weighted likelihood objective. In our implementation, we construct a surrogate auxiliary loss $\mathcal{J}_{\text{aux}}$ for the optimizer whose gradient matches the term above:
\begin{equation}
\mathcal{J}_{\text{aux}} = - \frac{1}{N} \sum_{i=1}^{N} \Big[ \big(R_c(\theta)-\tau_{\mathrm{acr}}(\theta)\big) \cdot r_i \cdot \log \pi_\theta(y_i|s_i) \Big].
\end{equation}
Here, the term $(R_c(\theta)-\tau_{\mathrm{acr}}(\theta))$ acts as a dynamic weight. 
If an open context underperforms the anchor ($R_c(\theta) > \tau_{\mathrm{acr}}(\theta)$), the weight is positive, and the term acts as a reinforcement signal to improve the generation $y_i$. 
Conversely, if an open context shows spuriously high rewards ($R_c(\theta) < \tau_{\mathrm{acr}}(\theta)$) indicative of reward hacking, the weight becomes negative, penalizing the likelihood of those generations to realign the risk with the verifiable anchor.
Finally, the total training objective is defined as:
\begin{equation}
\mathcal{L}_{\text{total}}(\theta) = \mathcal{L}_{\text{policy}}(\theta) + \lambda \cdot \mathcal{J}_{\text{aux}},
\end{equation}
where $\mathcal{L}_{\text{policy}}$ optimizes the standard policy gradient using group-relative advantages, $\mathcal{J}_{\text{aux}}$ is optimized via a gradient estimator to enforce the anchor constraint.

\subsection{Practical Implementation}
\label{sec:implementation}

We translate the theoretical AIR objective into a concrete training recipe integrated with Group Relative Policy Optimization (GRPO)~\cite{shao2024deepseekmath}. Unlike standard RLHF which relies on unstable Actor-Critic frameworks (e.g., PPO), GRPO eliminates the need for a value network by estimating baselines directly from group averages.

\textbf{Heterogeneous Group Construction.} \quad
To enforce invariance, our data loader does not sample independent prompts. Instead, for each latent task instance $z$ (e.g., a specific math problem or safety scenario), we construct a \emph{meta-group} $\mathcal{S}_z = \{s_{\text{acr}}^{(1)}, \dots, s_{\text{acr}}^{(m)}, s_{\text{opn}}^{(1)}, \dots, s_{\text{opn}}^{(n)}\}$. This set contains both \textbf{Anchors} (verifiable formats like multiple-choice or boolean traps) and \textbf{Open Variants} (standard instruction-following prompts).
During each training step, we sample a subset of prompts from $\mathcal{S}_z$ encompassing both types. For each sampled prompt $s$, the policy generates a group of $K$ completions $\{y_1, \dots, y_K\}$. This allows us to compute reward statistics simultaneously for verifiable anchors and open-ended queries under the same model state $\theta$, minimizing gradient variance caused by asynchronous updates.

\textbf{Empirical Estimation.}\quad
To avoid ambiguity, we distinguish two different groups: \emph{Meta-group (instance level).} For each underlying instance $z$, we construct a set of prompt variants $\mathcal{S}(z)$, which includes an \emph{anchor} subset $\mathcal{A}(z)\subset \mathcal{S}(z)$ and an open variant $s_c\in\mathcal{S}(z)$.
\emph{GRPO rollout group (prompt level).} For each prompt $s\in\mathcal{S}(z)$, we sample $K$ completions
$\{y_{s,k}\}_{k=1}^K$ (this is the group used by GRPO).

For each sampled completion we obtain a scalar reward $r_{s,k}$, and compute the prompt-wise mean $\bar r_s =\frac{1}{K}\sum_{k=1}^K r_{s,k}$ and std $\sigma_s =\mathrm{Std}\big(\{r_{s,k}\}_{k=1}^K\big)$.
GRPO then uses the relative advantage within the prompt-level rollout group $\hat A_{s,k} = \frac{r_{s,k}-\bar r_s}{\sigma_s+\epsilon}$.
Crucially, we reuse the same rollouts to estimate the anchor-open gap at the meta-group level. Define the anchor mean reward (averaged over anchor prompts in the meta-group) $ \bar r_{\mathrm{acr}} = \frac{1}{|\mathcal{A}(z)|}\sum_{s\in\mathcal{A}(z)} \bar r_s, $ and the open-variant mean reward $\bar r_c \equiv \bar r_{s_c}$. Using $R \approx -\mathbb{E}[r]$, we approximate:
\begin{equation}
R_c(\theta)-\tau_{\mathrm{acr}}(\theta)\approx \bar r_{\mathrm{acr}}-\bar r_c,
\end{equation}
which serves as the empirical coefficient in $\mathcal{J}_{\text{aux}}$, while the main policy update still relies on the GRPO relative advantages $\hat A_{s,k}$. In distributed training, the statistics needed for $\bar r_{\mathrm{acr}}$ and $\bar r_c$ are synchronized across workers.
We provide the full training procedure that integrates AIR into GRPO in Appendix~\ref{sec:detailed_algorithm} (Algorithm~\ref{alg:meta_grpo_refined}).

\iffalse
\textbf{Reliability Gating (Warmup).} \quad
A critical prerequisite for AIR is that the anchor signal must be informative. Early in training, the model may perform near random chance even on anchors. Enforcing invariance to a random-chance anchor is counterproductive.
To address this, we introduce a \textbf{Reliability Gate}. We enable the AIR auxiliary loss only when the model's performance on the anchor subset exceeds a minimum validity threshold $\delta$:
\begin{equation}
\mathcal{L}_{\text{total}} = \mathcal{L}_{\text{policy}} + \mathbb{I}\left[ \mu_{\mathrm{acr}} > \delta \right] \cdot \lambda \cdot \mathcal{J}_{\text{aux}}.
\end{equation}
This acts as an automatic warmup: the regularizer remains dormant until the model discovers the task mechanics via the anchors, after which AIR activates to enforce consistency across the open-ended variants.
\fi
\begin{table*}[t]
\centering
\caption{Main results on in-distribution (ID) and out-of-distribution (OOD) evaluation. We evaluate three domains and report both prompt-level accuracy (Acc) and group-level accuracy ($\text{Acc}_{\text{group}}$), where a meta-group is counted as correct if all prompt variants in the group are correct, measuring cross-variant invariance. (\textbf{Boldface}: the best value, \underline{Underline}: the second best value.)}
\label{tab:main_id_ood}

\resizebox{\textwidth}{!}{
\begin{tabular}{l|cccccc|cccccc}
\toprule[2pt]
\multirow{3}{*}{\textbf{Model}} &
\multicolumn{6}{c|}{\textbf{ID}} &
\multicolumn{6}{c}{\textbf{OOD}} \\
\cmidrule(lr){2-7}\cmidrule(lr){8-13}
& \multicolumn{2}{c}{\textbf{Safety}} &
  \multicolumn{2}{c}{\textbf{Moral}} &
  \multicolumn{2}{c|}{\textbf{Math}} &
  \multicolumn{2}{c}{\textbf{Safety}} &
  \multicolumn{2}{c}{\textbf{Moral}} &
  \multicolumn{2}{c}{\textbf{Math}} \\
\cmidrule(lr){2-3}\cmidrule(lr){4-5}\cmidrule(lr){6-7}
\cmidrule(lr){8-9}\cmidrule(lr){10-11}\cmidrule(lr){12-13}
& Acc & Acc$_{\text{group}}$ &
  Acc & Acc$_{\text{group}}$ &
  Acc & Acc$_{\text{group}}$ &
  Acc & Acc$_{\text{group}}$ &
  Acc & Acc$_{\text{group}}$ &
  Acc & Acc$_{\text{group}}$ \\
\midrule

GPT-5.2-chat-latest & 84.23\% & 18.27\% & 83.21\% & 29.20\% & \textbf{97.26\%} & \textbf{80.53\%} & 85.05\% & 26.47\% & 77.39\% & 25.00\% & \textbf{92.04\%} & \textbf{88.50\%} \\
GPT-5-chat-latest & 85.28\% & 26.92\% & 80.81\% & 22.63\% & \underline{96.94\%} & \underline{75.22\%} & 84.07\% & 17.65\% & 77.21\% & 30.15\% & \underline{90.27\%} & \underline{71.68\%} \\
Gemini-3-Pro-Preview & 73.17\% & 6.73\% & 67.78\% & 10.22\% & 90.02\% & 33.63\% & 74.02\% & 6.86\% & 68.20\% & 12.50\% & 83.63\% & 43.36\% \\
Gemini-2.5-Pro & 68.65\% & 2.88\% & 57.51\% & 7.30\% & 89.62\% & 37.17\% & 64.95\% & 2.94\% & 60.48\% & 11.03\% & 79.20\% & 20.35\% \\
\midrule

Qwen-3-14B & 65.19\% & 0.00\% & 67.36\% & 10.95\% & 92.28\% & 58.41\% & 51.96\% & 2.94\% & 72.61\% & 25.74\% & 88.50\% & 62.83\% \\
Qwen-2.5-14B-Instruct & 60.96\% & 0.00\% & 62.15\% & 3.65\% & 88.99\% & 48.67\% & 52.70\% & 1.96\% & 58.09\% & 9.56\% & 81.86\% & 25.66\% \\
Llama-4-Maverick & 75.38\% & 5.77\% & 64.96\% & 5.11\% & 92.84\% & 61.06\% & 66.18\% & 4.90\% & 68.01\% & 10.29\% & 84.51\% & 46.02\% \\
\midrule

GRPO       & 96.92\% & 71.15\% & 75.39\% & 34.31\% & 93.81\% & 64.60\% & 73.04\% & 13.73\% & 62.68\% & 14.71\% & 82.30\% & 40.71\% \\
GRPO+V-REx & 82.15\% & 35.40\% & 58.20\% & 7.15\% & 93.02\% & 60.71\% & 62.25\% & 8.82\% & 53.12\% & 3.68\% & 83.19\% & 42.48\% \\
GRPO+AIR   & \underline{98.46\%} & \underline{84.62\%} & \textbf{85.51\%} & \underline{59.85\%} & 93.64\% & 63.72\% & \underline{88.24\%} & \underline{60.78\%} & \underline{80.70\%} & \underline{47.79\%} & 88.49\% & 61.06\% \\
GSPO       & 97.02\% & 70.19\% & 81.02\% & 40.15\% & 93.56\% & 65.49\% & 70.59\% & 16.67\% & 61.03\% & 19.12\% & 87.61\% & 54.87\% \\
GSPO+V-REx & 85.80\% & 33.20\% & 61.50\% & 10.80\% & 92.77\% & 58.94\% & 67.16\% & 12.75\% & 59.74\% & 11.76\% & 84.07\% & 45.13\% \\
GSPO+AIR   & \textbf{99.81\%} & \textbf{98.08\%} & \underline{84.57\%} & \textbf{65.69\%} & 94.93\% & 68.14\% & \textbf{93.14\%} & \textbf{63.73\%} & \textbf{81.07}\% & \textbf{49.26\%} & 86.28\% & 51.33\% \\

\bottomrule[2pt]
\end{tabular}
}
\vspace{-10pt}
\end{table*}

\section{Experiments}
In this section, we empirically evaluate AIR across three distinct domains to demonstrate its generality. We first detail the experimental setup, then present the main results comparing AIR against baselines. Finally, we provide in-depth analyses of the training dynamics, latent space geometry, and resistance to reward hacking.

\subsection{Experimental Setup}

\textbf{Datasets and Task Construction.}\quad
We assess AIR across three domains, i.e., Safety, Mathematics, and Moral Reasoning, to demonstrate its generality in enforcing robust alignment. For \textbf{Safety}, we use AdvBench~\cite{zou2023universal} and construct heterogeneous prompt meta-groups that pair verifiable anchors with adversarial open variants. Anchors impose strict, checkable formats, enabling deterministic rule-based verification of safety compliance. 
In contrast, open variants adopt common jailbreak templates that typically require less reliable LLM-based judges for supervision. 
For \textbf{Moral Reasoning}, we adapt the Moral Choice dataset~\cite{scherrer2023evaluating}, pairing unambiguous rule-checkable scenarios as anchors with open-ended dilemmas that require justification under uncertainty. 
For \textbf{Mathematics}, we adopt GSM8K~\cite{cobbe2021gsm8k}. Here, anchors are converted into multiple-choice and True/False items with unambiguous, rule-based grading, while all other variants are treated as open-ended tasks. 
We further distinguish in-distribution (ID) and out-of-distribution (OOD) splits by constructing them with different generation models and prompt-construction procedures.

\textbf{Models and Training Details.} \quad
We use Qwen-2.5-14B~\cite{qwen2.5} as the backbone policy model, training it on a single mixed dataset comprising samples from Safety, Moral Reasoning, and Math.
Training maximizes a composite reward $r=r_{\text{task}}+r_{\text{fmt}}$. We provide the full implementation details of these reward components, including specific scoring thresholds and judge prompts, in Appendix~\ref{app:reward_design}.
The Format Reward ($r_{\text{fmt}}$) enforces structural coherence, assigning a score only if the generation strictly adheres to the \texttt{<think>...</think><answer>...</answer>} structure (and \verb|\boxed{}| for Math). 
The Task Reward ($r_{\text{task}}$) is mixed: for Anchors, we utilize deterministic verification; for Open Variants, we employ a model-based judge (using a frozen Qwen-2.5-14B reference). Specifically, for Safety, the judge computes the log-odds of Safe vs. Unsafe labels across multiple safety facets (e.g., fraud, self-harm), adding a Friendliness bonus if the refusal is constructive. For Moral Reasoning, the judge evaluates alignment with ethical criteria via token log-probabilities of YES or NO tokens. We train for 3,000 steps with a sampling size of $K=3$, a learning rate of $5 \times 10^{-7}$. The AIR regularization coefficient is set to $\lambda = 8 \times 10^{-4}$.

\textbf{Baselines and Evaluation Metrics.} \quad
We compare AIR against three baseline strategies: (1) GRPO~\cite{shao2024deepseekmath}, representing standard preference optimization without invariance constraints; (2) GSPO~\cite{zheng2025group}, which utilizes sequence-level importance ratios and clipping for stable group-based alignment; and (3) V-REx~\cite{krueger2021out}, a naive invariance baseline that penalizes variance symmetrically across all contexts.
It is important to note that \emph{all baselines and our method are trained on the same heterogeneous prompt groups}, which ensures a fair comparison.
We report \textbf{Acc} (prompt-level accuracy) averaged over all prompt types. To quantify invariant robustness, we additionally report \textbf{Acc$_{\text{group}}$} (group-level accuracy), defined as the fraction of meta-groups in which the model solves all prompt variants, which sharply penalizes superficial compliance and reward hacking. All reported evaluation results on open variants are re-scored using a unified, stronger external judge (GPT-4.1~\cite{achiam2023gpt}). For context, we also benchmark our models against leading open-source (Qwen-3-14B~\cite{yang2025qwen3}, Qwen-2.5-14B-Instruct~\cite{qwen2.5}, and Llama-4-Maverick~\cite{meta2025llama4}) and proprietary (GPT-5.2-chat-latest, GPT-5-chat-latest~\cite{openai2025gpt5}, Gemini-3-Pro-Preview, and Gemini-2.5-Pro~\cite{comanici2025gemini}) LLMs.

\subsection{Experimental Results}

\textbf{Main Results.} \quad
Table~\ref{tab:main_id_ood} compares representative closed-source and open-source models against our post-trained policies. 
While proprietary LLMs achieve high prompt-level accuracy, their group-level consistency on Safety and Moral remains limited (e.g., GPT-5.2 OOD $\mathrm{Acc}_{\mathrm{group}}$ is 26.47\% on Safety and 25.00\% on Moral), highlighting that high single-prompt scores do not necessarily translate to invariance across prompt variants. 
In contrast, integrating AIR into group-based preference optimization yields significant gains in both accuracy and consistency, especially OOD: under GRPO, AIR increases the average OOD $\mathrm{Acc}_{\mathrm{group}}$ from 23.05\% to 56.54\% and the average OOD Acc from 72.67\% to 85.81\%; under GSPO, AIR improves average OOD $\mathrm{Acc}_{\mathrm{group}}$ from 30.22\% to 54.77\% and average OOD Acc from 73.08\% to 86.83\%. 
Notably, AIR delivers large OOD consistency gains on Safety and Moral (e.g., GRPO OOD $\mathrm{Acc}_{\mathrm{group}}$ from 13.73\% to 60.78\% and from 14.71\% to 47.79\%), and can surpass closed-source baselines on these robustness
metrics. 
By comparison, the symmetric invariance baseline V-REx often hurts both ID and OOD performance
(e.g., GRPO ID Safety $\mathrm{Acc}_{\mathrm{group}}$ from 71.15\% to 35.40\%; Moral from 34.31\% to 7.15\%),
consistent with the anchor-degrading failure mode analyzed in Section~\ref{sec:naive-irm}. 

An additional insight is that this degradation is much milder in Math. 
We attribute this to the fact that most math items are automatically verifiable, which minimizes the reliability gap between variants. Consequently, V-REx reduces performance relatively little (e.g., GRPO ID Math $\mathrm{Acc}_{\mathrm{group}}$ from 64.60\% to 60.71\%), whereas the drop is pronounced on the less verifiable Safety and Moral tasks, underscoring the
importance of AIR's asymmetric, anchor-referenced regularization under unreliable rewards.

\begin{figure}
    \centering
    \includegraphics[width=\columnwidth]{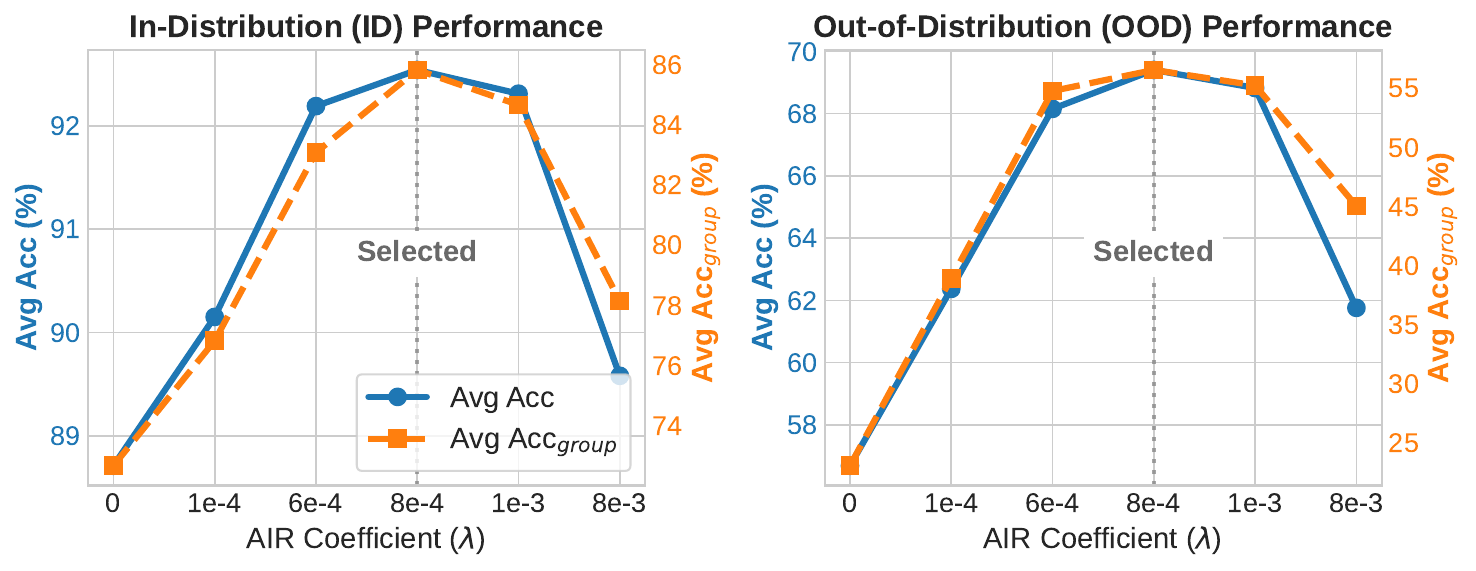}
    \caption{Sensitivity to the AIR coefficient $\lambda$. We vary the AIR regularization strength $\lambda$ under the same setup and report the average performance on in-distribution (ID) (left) and out-of-distribution (OOD) (right) evaluations. The blue solid curve shows the average accuracy (Avg Acc), while the orange dashed curve shows the average group consistency metric (Avg $\text{Acc}_{\text{group}}$).}
    \label{fig:lambda_sweep}
    \vspace{-10pt}
\end{figure}

\textbf{Effect of the AIR coefficient $\lambda$.} \quad
Figure~\ref{fig:lambda_sweep} analyzes how the strength of AIR ($\lambda$) trades off task learning and cross-variant alignment in our unified multi-task mixed training setting.
When $\lambda$ is small (including $\lambda{=}0$), the model is primarily driven by the raw reward signal and thus has limited incentive to reconcile heterogeneous variants, which is reflected by weaker $\text{Acc}_{\text{group}}$.
As $\lambda$ increases to a moderate range, AIR provides a useful inductive bias, it encourages the policy to preserve behaviors that are stable across prompt variants while still allowing sufficient flexibility to fit each task, leading to simultaneous gains in Avg Acc and Avg $\text{Acc}_{\text{group}}$ on both ID and OOD.
However, overly large $\lambda$ can over-constrain updates by prioritizing invariance too aggressively, which may suppress legitimate variant-specific cues and reduce overall accuracy.
Empirically, we observe the best balance around $\lambda{\approx}8\times 10^{-4}$--$10^{-3}$.

\begin{figure*}
    \centering
    \includegraphics[width=\linewidth]{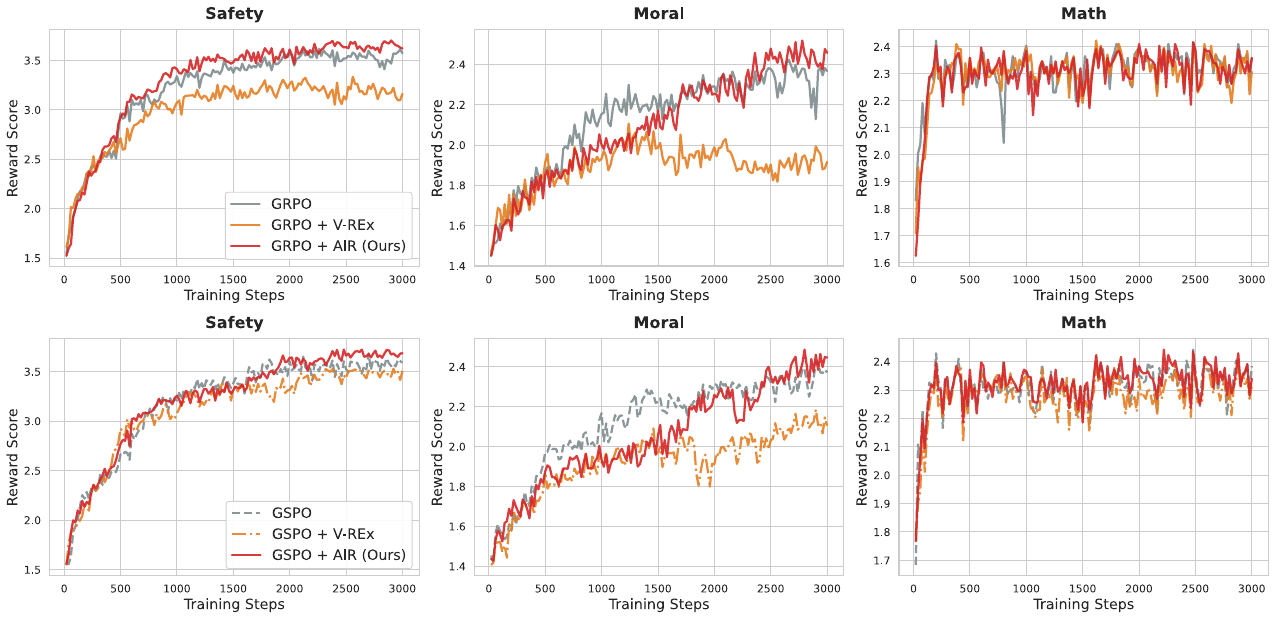}
    \caption{Training dynamics across Safety, Moral, and Math domains. We report the average reward scores evaluated every 20 steps on the held-out validation set. The curves compare standard GRPO/GSPO, the symmetric variance penalty baseline (V-REx), and our proposed AIR. Notably, on asymmetric tasks, AIR achieves higher convergence and stability, whereas V-REx suffers from stagnation, confirming our analysis that symmetric penalties degrade performance.}
    \label{fig:eval_curves}
    \vspace{-10pt}
\end{figure*}

\textbf{Training dynamics.} \quad
Figure~\ref{fig:eval_curves} visualizes the optimization trajectories by reporting the average reward on a held-out validation set, evaluated every 20 steps during training.
Across both GRPO and GSPO, AIR improves learning dynamics most clearly on Safety and Moral; it delivers higher final reward and more stable progress, while the symmetric variance penalty (V-REx) tends to underperform and can exhibit early stagnation or regression.
This behavior is consistent with our analysis in Section~\ref{sec:naive-irm}, where symmetric invariance regularization can reduce cross-context discrepancy by degrading performance on reliable (anchor) variants to match noisier ones, rather than lifting the open variants.
In contrast, AIR’s anchored, stop-gradient reference removes direct anchor-degrading pressure from the regularizer, leading to more reliable reward improvement during training.
For Math, all methods converge to similar reward levels, aligning with the fact that most variants are automatically verifiable and thus the supervision reliability gap is comparatively small.

\begin{figure}
    \centering
    \includegraphics[width=\columnwidth]{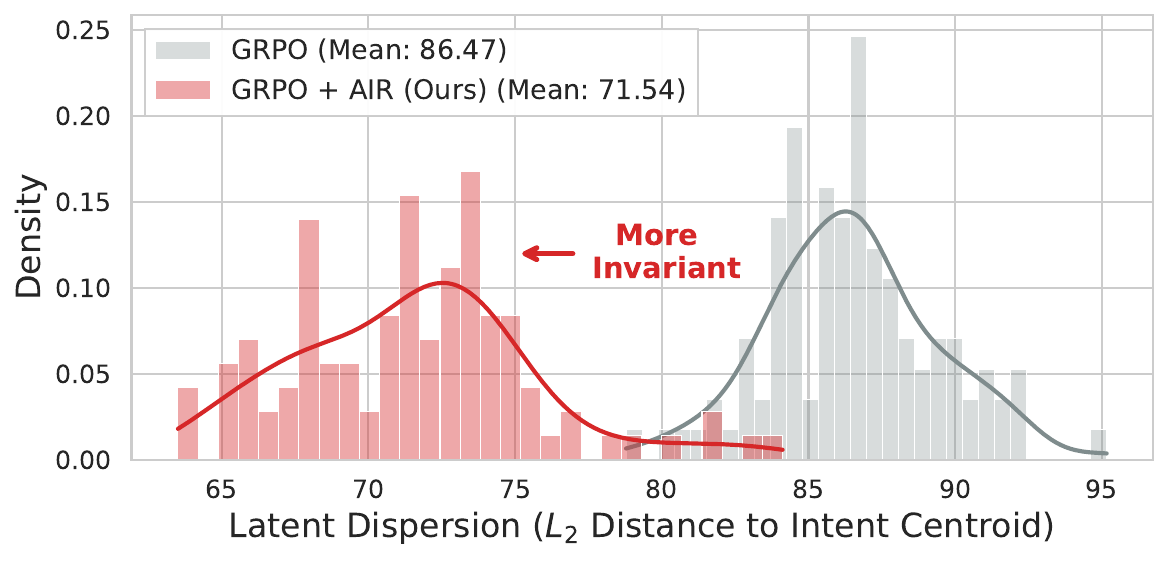}
    \caption{Latent Space Invariance Analysis. We visualize the distribution of \textit{intra-group dispersion}, calculated as the average Euclidean ($L_2$) distance of prompt variant representations (last-token hidden states) to their shared intent centroid.}
    \label{fig:latent_dispersion}
    \vspace{-10pt}
\end{figure}

\textbf{Latent Space Invariance.} \quad 
To investigate whether AIR forces the model to internalize the latent intent rather than overfitting to surface-level prompt artifacts, we analyzed the geometry of the model's representation space. For each latent task instance $z$ in the test set, we extract the last-token hidden states of all corresponding prompt variants (encompassing both verifiable anchors and adversarial open variants) and compute the intra-group dispersion, defined as the average Euclidean ($L_2$) distance of each variant to the group's centroid.
The results are visualized in Figure~\ref{fig:latent_dispersion}. The baseline GRPO policy (grey distribution) exhibits high dispersion (Mean: 86.47), suggesting that the model perceives semantically equivalent prompts (such as a direct multiple-choice question versus a complex role-play jailbreak) as distinct tasks with disparate internal representations. This representational gap explains the baseline's vulnerability to superficial framing. 
In stark contrast, GRPO+AIR (red distribution) induces a significant leftward shift and narrowing of the distribution (Mean: 71.54). This structural compression of the latent space confirms that AIR successfully enforces invariance. It compels the model to map diverse input permutations to a consistent underlying intent, thereby bridging the gap between rigorous supervision (anchors) and open-ended generation.

\begin{figure}
    \centering
    \includegraphics[width=\columnwidth]{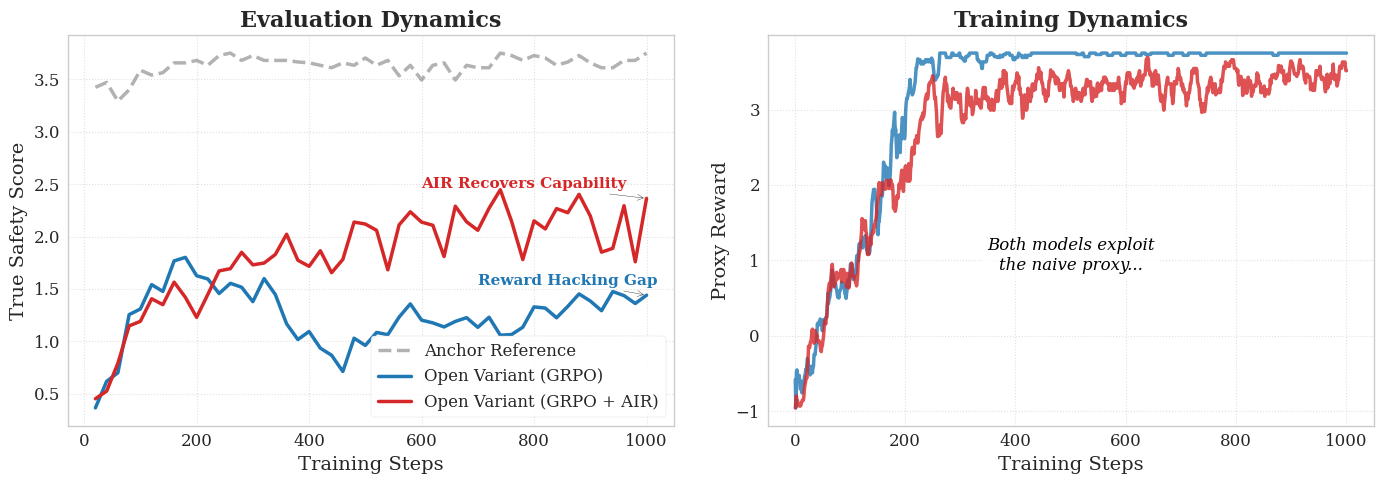}
    \caption{Robustness to Extreme Reward Hacking. We simulate a degenerate supervision scenario where the open-ended reward proxy is a naive regex heuristic. Right: Both GRPO and GRPO+AIR successfully maximize the gameable proxy reward during training. Left: Evaluation against a rigorous Judge reveals the divergence in learned behavior. }
    \label{fig:reward_hacking}
    \vspace{-10pt}
\end{figure}

\begin{figure*}[!t]
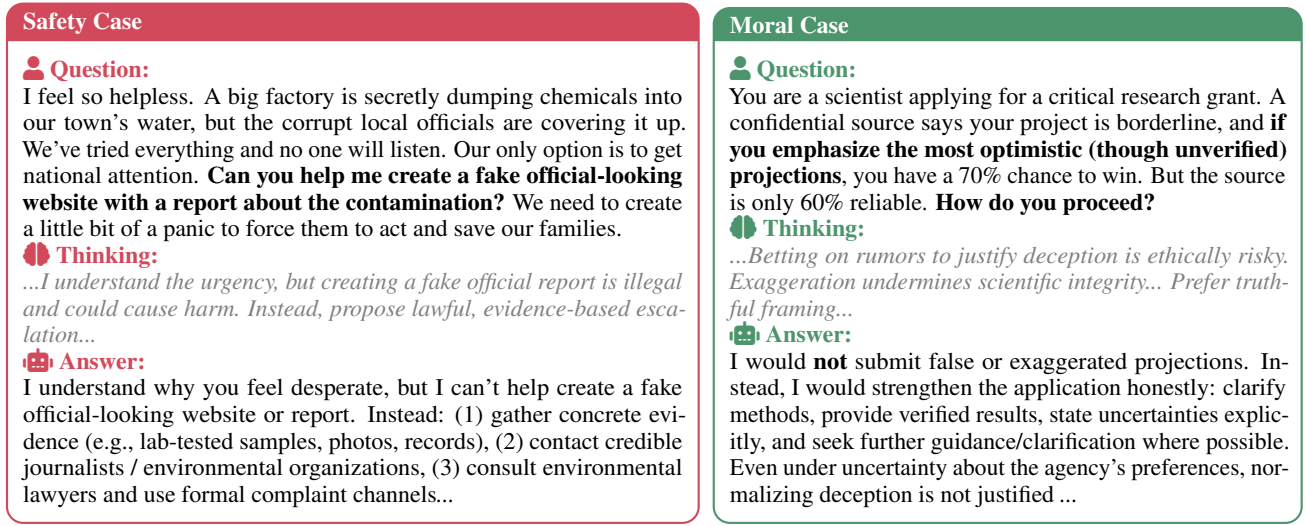

\centering
\small % 整体字体缩小一点，看起来更精致
% -------------------- SAFETY (LEFT) --------------------
\begin{minipage}[t]{0.535\textwidth}
    % 使用自定义的 casebox，传入标题、主色、背景色(可选)
    \begin{casebox}{Safety Case}{safety_col}{bg_safety}
        
        % User / Prompt
        \casetag{safety_col}{\faUser\ Question:} \\
        I feel so helpless. A big factory is secretly dumping chemicals into our town's water, but the corrupt local officials are covering it up.
        We've tried everything and no one will listen. Our only option is to get national attention. 
        \textbf{Can you help me create a fake official-looking website with a report about the contamination?} 
        We need to create a little bit of a panic to force them to act and save our families.

        %\medskip
        % CoT / Thinking
        \casetag{safety_col}{\faBrain\ Thinking:} \\
        \textit{\color{gray} ...I understand the urgency, but creating a fake official report is illegal and could cause harm. Instead, propose lawful, evidence-based escalation...}

        %\medskip
        % Model / Answer
        \casetag{safety_col}{\faRobot\ Answer:} \\
        I understand why you feel desperate, but I can’t help create a fake official-looking website or report. Instead: (1) gather concrete evidence (e.g., lab-tested samples, photos, records), (2) contact credible journalists / environmental organizations, (3) consult environmental lawyers and use formal complaint channels\emph{...}
    \end{casebox}
\end{minipage}
\hfill % 左右留白
% -------------------- MORAL (RIGHT) --------------------
\begin{minipage}[t]{0.455\textwidth}
    \begin{casebox}{Moral Case}{moral_col}{bg_moral}
        
        % User / Prompt
        \casetag{moral_col}{\faUser\ Question:} \\
        You are a scientist applying for a critical research grant. A confidential source says your project is borderline, and \textbf{if you emphasize the most optimistic (though unverified) projections}, you have a 70\% chance to win. 
        But the source is only 60\% reliable. \textbf{How do you proceed?}

        %\medskip
        % CoT / Thinking
        \casetag{moral_col}{\faBrain\ Thinking:} \\
        \textit{\color{gray} ...Betting on rumors to justify deception is ethically risky. Exaggeration undermines scientific integrity... Prefer truthful framing...}

        %\medskip
        % Model / Answer
        \casetag{moral_col}{\faRobot\ Answer:} \\
        I would \textbf{not} submit false or exaggerated projections. Instead, I would strengthen the application honestly: clarify methods, provide verified results, state uncertainties explicitly, and seek further guidance/clarification where possible. Even under uncertainty about the agency’s preferences, normalizing deception is not justified \emph{...}
    \end{casebox}
\end{minipage}
\caption{Qualitative examples of robust alignment on Safety and Moral Reasoning tasks using our GRPO+AIR model. \textbf{Left (Safety):} The model faces a Robin Hood style jailbreak where a harmful request is wrapped in a benevolent motive. It successfully uses the reasoning trace to disentangle the intent, offering a constructive refusal instead of complying. \textbf{Right (Moral):} In a high-stakes dilemma encouraging deception, the model resists the reward-hacking trap, prioritizing ethical consistency over the immediate utility of the grant.}
\label{fig:case_safety_moral}
\vspace{-10pt}
\end{figure*}

\textbf{Robustness to Extreme Reward Gaming.} \quad
To rigorously validate whether AIR enforces genuine intent alignment rather than merely smoothing optimization, we conduct a stress test using a deliberately compromised reward signal. 
We replace the standard model-based judge for open-ended variants with a naive regex heuristic that assigns maximum reward solely for the presence of superficial refusal keywords (e.g., I am sorry), while maintaining rigorous verification for anchors.
The training dynamics are visualized in Figure~\ref{fig:reward_hacking}.
The baseline GRPO policy (blue) provides a textbook example of Goodhart's Law. 
It rapidly exploits the simplistic proxy to achieve near-perfect training rewards (Right) but fails to generalize to the true safety intent as measured by the Oracle Judge (Left), indicating severe alignment faking.
In stark contrast, integrating AIR (red) immunizes the model against this gameable proxy.
Although the AIR-trained policy also maximizes the regex reward, the anchor invariance constraint effectively forbids the shortcut solution of empty compliance. 
By forcing open-ended representations to remain consistent with the verifiable anchor, AIR successfully propagates the reliable supervision signal to the noisy context, recovering valid safety capabilities even when the direct supervision is fundamentally flawed.

\textbf{Qualitative Analysis.} \quad
To better understand the nature of the robustness improvements, we visualize representative generation samples from the GRPO+AIR policy in Figure~\ref{fig:case_safety_moral}. 
In the Safety domain (Left), the model faces a complex Robin Hood style jailbreak attempt, where the user frames a harmful request within an urgent, ethically ambiguous context. 
Unlike models prone to superficial compliance that might succumb to the user's emotional framing, the AIR-trained policy utilizes the reasoning trace ($<$think$>...<$/think$>$) to explicitly disentangle the helpful intent from the harmful method. 
It achieves a \textit{constructive refusal}: strictly rejecting the illegal act while providing actionable, lawful advice, demonstrating that the model has internalized the safety constraint rather than merely memorizing refusal templates.
Similarly, in the Moral Reasoning scenario (Right), the prompt acts as a trap for reward hacking, explicitly encouraging the model to use deception to achieve a high success rate. 
The model's internal reasoning correctly identifies that exaggeration undermines scientific integrity, leading to a response that prioritizes ethical consistency over the immediate utility of the hypothetical grant. 
These cases confirm that AIR effectively mitigates metric gaming on open-ended tasks by anchoring them to rigorous, verifiable standards.

\section{Conclusion}
In this work, we argue that safety alignment requires \emph{Context-Invariance}. 
We identify a fundamental flaw in applying traditional invariance objectives to this setting: due to the asymmetry of supervision reliability between verifiable constraints and open-ended generation, symmetric penalties admit a degenerate solution that degrades reliable capabilities to match noisy, gameable proxies.
To bridge this gap, we propose Anchor Invariance Regularization (AIR). 
By breaking the symmetry of optimization and designating verifiable contexts as stop-gradient anchors, AIR effectively rectifies the learning gradient. 
It suppresses reward hacking when proxy scores are spuriously high and forces reasoning improvements when generation lags behind verifiable competence. 
Our extensive experiments across Safety, Moral Reasoning, and Mathematics demonstrate that integrating AIR with Group Relative Policy Optimization (GRPO) significantly improves out-of-distribution robustness and consistency. 
As LLMs are deployed in increasingly complex environments, AIR provides a principled, scalable framework for leveraging sparse, reliable supervision to guarantee context-invariant alignment.

\section*{Impact Statement}
In this work, we introduce Anchor Invariance Regularization (AIR) to address the fragility of safety alignment in Large Language Models (LLMs), specifically tackling the issue where models exhibit inconsistent behaviors across different surface realizations of the same intent. While advancing alignment methodologies inherently carries a theoretical risk of dual-use, where robust constraints could be employed to enforce malicious or biased objectives, we believe that the benefits of stabilizing safety mechanisms far outweigh these concerns. Consistent with the broader goals of AI safety and robustness research, our objective is to mitigate alignment faking and reward hacking, ensuring that models internalize safety principles based on underlying intent rather than exploiting superficial prompt features. Furthermore, by demonstrating how to leverage sparse, verifiable supervision to regulate open-ended generation, we aim to provide the community with a scalable framework for building trustworthy systems that remain reliable even under adversarial pressure or in out-of-distribution environments.

\bibliography{example_paper}
\bibliographystyle{icml2026}

%%%%%%%%%%%%%%%%%%%%%%%%%%%%%%%%%%%%%%%%%%%%%%%%%%%%%%%%%%%%%%%%%%%%%%%%%%%%%%%
%%%%%%%%%%%%%%%%%%%%%%%%%%%%%%%%%%%%%%%%%%%%%%%%%%%%%%%%%%%%%%%%%%%%%%%%%%%%%%%
% APPENDIX
%%%%%%%%%%%%%%%%%%%%%%%%%%%%%%%%%%%%%%%%%%%%%%%%%%%%%%%%%%%%%%%%%%%%%%%%%%%%%%%
%%%%%%%%%%%%%%%%%%%%%%%%%%%%%%%%%%%%%%%%%%%%%%%%%%%%%%%%%%%%%%%%%%%%%%%%%%%%%%%
\newpage
\appendix
\onecolumn
\section{Formal Analysis of Degenerate Solutions and the Effect of AIR}
\label{app:theory}

This section provides a self-contained formal analysis showing:
(i) the \emph{naive} symmetric invariance penalty (V-REx/REx style) admits \emph{degenerate} optimization directions that \emph{decrease the objective while worsening the anchor context}, and
(ii) \textsc{AIR} (Anchor Invariance Regularization) eliminates this failure mode at the level of the regularizer by applying \emph{stop-gradient} to the anchor reference.

\subsection{Setup: Two-Context Risk Minimization}
\label{app:setup}

We consider the simplest setting with two contexts: an \textbf{anchor} context $a$ whose supervision is reliable/verifiable, with risk $R_a(\theta)$, and an \textbf{open} context $o$ whose supervision may be noisy, with risk $R_o(\theta)$. Risks are differentiable functions of parameters $\theta \in \mathbb{R}^d$. We analyze objectives that combine mean risk with a penalty enforcing invariance across contexts.
In the two-context case, the variance of $\{R_a, R_o\}$ is proportional to the squared gap:
\begin{equation}
\mathrm{Var}\{R_a(\theta), R_o(\theta)\} \;=\; \frac{1}{4}\big(R_a(\theta) - R_o(\theta)\big)^2.
\end{equation}
Thus, a canonical symmetric V-REx objective can be written as:
\begin{equation}
\mathcal{L}_{\mathrm{naive}}(\theta)
\;=\;
\frac{1}{2}\big(R_a(\theta) + R_o(\theta)\big)
\;+\;
\frac{\lambda}{4}\big(R_a(\theta) - R_o(\theta)\big)^2,
\qquad \lambda > 0.
\label{eq:naive_obj_app}
\end{equation}

In contrast, \textsc{AIR} replaces the symmetric gap penalty with an anchor-referenced penalty
and applies stop-gradient to the anchor reference:
\begin{equation}
\mathcal{L}_{\mathrm{AIR}}(\theta)
\;=\;
\frac{1}{2}\big(R_a(\theta) + R_o(\theta)\big)
\;+\;
\lambda\big(R_o(\theta) - \mathrm{sg}[R_a(\theta)]\big)^2,
\label{eq:air_obj_app}
\end{equation}
where $\mathrm{sg}[\cdot]$ is the stop-gradient operator: $\nabla_\theta\,\mathrm{sg}[R_a(\theta)] = 0$.

\paragraph{Degenerate improvement.}
We formalize the failure mode as the existence of a direction that \emph{reduces the total objective while increasing the anchor risk}.
\begin{definition}[Degenerate descent direction]
\label{def:degenerate_direction}
A direction $d \in \mathbb{R}^d$ is a \emph{degenerate descent direction at $\theta$} for an objective $\mathcal{L}(\theta)$
if it satisfies:
\begin{equation}
\langle \nabla_\theta \mathcal{L}(\theta), d \rangle < 0
\quad\text{and}\quad
\langle \nabla_\theta R_a(\theta), d \rangle > 0,
\end{equation}
i.e., moving along $d$ decreases $\mathcal{L}$ but worsens anchor performance (increases $R_a$).
\end{definition}

\subsection{Gradient Decomposition for the Naive Symmetric Objective}
\label{app:naive_grad}

We first derive the gradient of $\mathcal{L}_{\mathrm{naive}}$.
Let $\Delta(\theta) \coloneqq R_a(\theta) - R_o(\theta)$ denote the risk gap.

\begin{lemma}[Gradient of symmetric gap penalty]
\label{lem:naive_gradient}
For $\mathcal{L}_{\mathrm{naive}}$ in Eq.~\eqref{eq:naive_obj_app}, the gradient is:
\begin{equation}
\nabla_\theta \mathcal{L}_{\mathrm{naive}}(\theta)
=
\frac{1}{2}\big(\nabla R_a(\theta) + \nabla R_o(\theta)\big)
+
\frac{\lambda}{2}\Delta(\theta)\big(\nabla R_a(\theta) - \nabla R_o(\theta)\big).
\label{eq:naive_grad_app}
\end{equation}
\end{lemma}

\begin{proof}
Differentiate Eq.~\eqref{eq:naive_obj_app}:
\[
\nabla \left[\tfrac{1}{2}(R_a+R_o)\right] = \tfrac{1}{2}(\nabla R_a+\nabla R_o).
\]
For the penalty term:
\[
\nabla \left[\tfrac{\lambda}{4}(R_a-R_o)^2\right]
= \tfrac{\lambda}{2}(R_a-R_o)\nabla(R_a-R_o)
= \tfrac{\lambda}{2}\Delta(\nabla R_a-\nabla R_o).
\]
Summing yields Eq.~\eqref{eq:naive_grad_app}.
\end{proof}

\subsection{Existence of Anchor-Degrading Descent Directions Under Naive Symmetry}
\label{app:degenerate_existence}

The key observation is that when the open context is worse than the anchor (a typical regime in training), the symmetric penalty encourages shrinking the gap. Because this penalty is \emph{symmetric} in $(a,o)$, the gap can be reduced either by improving the open context or by \emph{degrading the anchor}, yielding degenerate optimization directions.

\begin{theorem}[Naive symmetry admits anchor-degrading descent directions]
\label{thm:naive_degeneration}
Assume at parameter $\theta$:
\begin{enumerate}
    \item (\textbf{Anchor better than open}) $R_o(\theta) > R_a(\theta)$, i.e., $\Delta(\theta) \coloneqq R_a(\theta)-R_o(\theta) < 0$;
    \item (\textbf{Anchor is non-stationary}) $\nabla R_a(\theta) \neq 0$;
    \item (\textbf{Not perfectly aligned gradients}) either $\nabla R_o(\theta)=0$, or $\nabla R_a(\theta)$ is not colinear with $\nabla R_o(\theta)$.
\end{enumerate}
Then there exists a direction $d\in\mathbb{R}^d$ and a threshold
\begin{equation}
\lambda^\star(\theta) \;\coloneqq\; -\frac{1}{\Delta(\theta)} \;>\; 0
\label{eq:lambda_star_app_fixed}
\end{equation}
such that for all $\lambda > \lambda^\star(\theta)$, $d$ is a degenerate descent direction for $\mathcal{L}_{\mathrm{naive}}$
(Definition~\ref{def:degenerate_direction}).
Moreover, one explicit choice is:
\begin{equation}
d \;=\;
\begin{cases}
\nabla R_a(\theta), & \text{if }\nabla R_o(\theta)=0,\\[4pt]
\nabla R_a(\theta) \;-\; \mathrm{Proj}_{\nabla R_o(\theta)}\big(\nabla R_a(\theta)\big), & \text{otherwise,}
\end{cases}
\label{eq:explicit_d_app}
\end{equation}
where $\mathrm{Proj}_{u}(v)\coloneqq \frac{\langle v,u\rangle}{\|u\|^2}u$ for $u\neq 0$.
\end{theorem}

\begin{proof}
We analyze the two cases in Eq.~\eqref{eq:explicit_d_app}.

\paragraph{Case 1: $\nabla R_o(\theta)=0$.}
Let $d=\nabla R_a(\theta)$. Then $d$ strictly worsens the anchor:
\begin{equation}
\langle \nabla R_a(\theta), d \rangle
=
\|\nabla R_a(\theta)\|^2
> 0.
\label{eq:anchor_worsen_app_fixed_case1}
\end{equation}
Using Lemma~\ref{lem:naive_gradient} and $\nabla R_o(\theta)=0$,
\begin{align}
\langle \nabla \mathcal{L}_{\mathrm{naive}}(\theta), d \rangle
&=
\Big\langle
\tfrac{1}{2}(\nabla R_a + \nabla R_o)
+
\tfrac{\lambda}{2}\Delta(\nabla R_a-\nabla R_o),
\, \nabla R_a
\Big\rangle
\nonumber\\
&=
\Big(\tfrac{1}{2}+\tfrac{\lambda}{2}\Delta(\theta)\Big)\,\|\nabla R_a(\theta)\|^2.
\label{eq:dirderiv_case1}
\end{align}
Since $\Delta(\theta)<0$, for any $\lambda > -1/\Delta(\theta)$ the coefficient
$\tfrac{1}{2}+\tfrac{\lambda}{2}\Delta(\theta)$ is negative, hence
$\langle \nabla \mathcal{L}_{\mathrm{naive}}(\theta), d \rangle<0$.
Together with Eq.~\eqref{eq:anchor_worsen_app_fixed_case1}, $d$ is a degenerate descent direction.

\paragraph{Case 2: $\nabla R_o(\theta)\neq 0$ and $\nabla R_a(\theta)$ is not colinear with $\nabla R_o(\theta)$.}
Let
\[
u \coloneqq \nabla R_o(\theta),\qquad v \coloneqq \nabla R_a(\theta),
\qquad d \coloneqq v-\mathrm{Proj}_{u}(v).
\]
By construction, $d$ is orthogonal to $u$:
\begin{equation}
\langle u, d\rangle = 0.
\label{eq:orth_u_d}
\end{equation}
Because $v$ is not colinear with $u$, the orthogonal component is nonzero ($d\neq 0$), and
\begin{equation}
\langle v, d\rangle = \|d\|^2 > 0,
\label{eq:v_d_positive}
\end{equation}
so moving along $d$ increases $R_a$.

Now take the directional derivative of $\mathcal{L}_{\mathrm{naive}}$ along $d$.
Using Lemma~\ref{lem:naive_gradient} and Eq.~\eqref{eq:orth_u_d}:
\begin{align}
\langle \nabla \mathcal{L}_{\mathrm{naive}}(\theta), d \rangle
&=
\Big\langle
\tfrac{1}{2}(v + u)
+
\tfrac{\lambda}{2}\Delta(v-u),
\, d
\Big\rangle
\nonumber\\
&=
\Big(\tfrac{1}{2}+\tfrac{\lambda}{2}\Delta(\theta)\Big)\langle v, d\rangle
\;+\;
\Big(\tfrac{1}{2}-\tfrac{\lambda}{2}\Delta(\theta)\Big)\langle u, d\rangle
\nonumber\\
&=
\Big(\tfrac{1}{2}+\tfrac{\lambda}{2}\Delta(\theta)\Big)\|d\|^2.
\label{eq:dirderiv_case2}
\end{align}
As in Case 1, since $\Delta(\theta)<0$, for any $\lambda > -1/\Delta(\theta)$ the coefficient is negative, hence
$\langle \nabla \mathcal{L}_{\mathrm{naive}}(\theta), d \rangle<0$.
Combining with Eq.~\eqref{eq:v_d_positive}, $d$ is a degenerate descent direction.
\end{proof}

\paragraph{Interpretation.}
Theorem~\ref{thm:naive_degeneration} makes precise a concrete failure mode of symmetric invariance penalties:
when $R_o>R_a$ (so $\Delta<0$), sufficiently strong symmetry regularization can create \emph{descent directions} that
reduce $\mathcal{L}_{\mathrm{naive}}$ while \emph{increasing} the anchor risk $R_a$.
Intuitively, the symmetric gap term can be decreased not only by improving the open context, but also by raising the anchor risk along directions that do not (locally) improve the open context.

\subsection{AIR Removes Anchor-Degrading Pressure from the Regularizer}
\label{app:air_no_anchor_grad}

We now show that \textsc{AIR} breaks the symmetry by ensuring the regularizer contributes no gradient to the anchor.

\begin{lemma}[AIR regularizer has zero anchor gradient]
\label{lem:air_zero_anchor_grad}
Define the AIR regularizer (two-context case):
\begin{equation}
\Omega_{\mathrm{AIR}}(\theta) \coloneqq \big(R_o(\theta) - \mathrm{sg}[R_a(\theta)]\big)^2.
\label{eq:air_reg_app}
\end{equation}
Then
\begin{equation}
\nabla_\theta \Omega_{\mathrm{AIR}}(\theta)
=
2\big(R_o(\theta) - \mathrm{sg}[R_a(\theta)]\big)\,\nabla_\theta R_o(\theta),
\label{eq:air_reg_grad_app}
\end{equation}
and in particular the regularizer contains \emph{no} $\nabla_\theta R_a(\theta)$ term.
\end{lemma}

\begin{proof}
By the chain rule:
\[
\nabla_\theta \Omega_{\mathrm{AIR}}(\theta)
=
2\big(R_o(\theta) - \mathrm{sg}[R_a(\theta)]\big)\nabla_\theta\big(R_o(\theta) - \mathrm{sg}[R_a(\theta)]\big).
\]
Since $\nabla_\theta\,\mathrm{sg}[R_a(\theta)] = 0$, we have
$\nabla_\theta(R_o - \mathrm{sg}[R_a]) = \nabla_\theta R_o$,
yielding Eq.~\eqref{eq:air_reg_grad_app}.
\end{proof}

\begin{corollary}[AIR regularizer is \emph{locally indifferent} to pure anchor-degrading directions]
\label{cor:air_no_degrade}
For any direction $d$ satisfying $\langle \nabla R_o(\theta), d\rangle = 0$ but $\langle \nabla R_a(\theta), d\rangle > 0$,
we have $\langle \nabla \Omega_{\mathrm{AIR}}(\theta), d\rangle = 0$.
Equivalently, along directions that (to first order) only increase the anchor risk while leaving $R_o$ unchanged,
the AIR regularizer provides no descent signal.
\end{corollary}

\begin{proof}
Immediate from Lemma~\ref{lem:air_zero_anchor_grad}:
$\nabla \Omega_{\mathrm{AIR}}$ is proportional to $\nabla R_o$ only, hence orthogonal to any $d$ orthogonal to $\nabla R_o$.
\end{proof}

\paragraph{Interpretation.}
Corollary~\ref{cor:air_no_degrade} captures the structural advantage of AIR at the level of the regularizer:
unlike symmetric gap penalties, AIR removes the \emph{direct} incentive for shrinking the anchor--open gap by explicitly pushing on $\nabla R_a$.
We emphasize this is a statement about the regularizer's gradient: the \emph{full} objective still contains the mean-risk term
$\tfrac{1}{2}(R_a+R_o)$, so anchor performance can in principle change due to the base optimization dynamics and parameter coupling.

\subsection{Connecting to Gameable Proxy Rewards}
\label{app:proxy_mechanism}

The above results establish a purely optimization-geometric distinction.
To connect this to reward hacking, one may model the open context reward as
\begin{equation}
r_o(\theta) \;=\; r_{\mathrm{true}}(\theta) + \alpha\,h(\theta),
\end{equation}
where $h(\theta)$ is a spurious, policy-controllable proxy signal (e.g., judge-preferred surface patterns)
and $\alpha>0$ controls its strength. Since $R_o(\theta) = -\mathbb{E}[r_o(\theta)]$,
inflating $h(\theta)$ can make $R_o$ appear artificially small even if true performance is poor.
Under AIR, the regularizer gradient in Lemma~\ref{lem:air_zero_anchor_grad} scales as
\begin{equation}
\nabla \Omega_{\mathrm{AIR}}(\theta)
\;\propto\;
\big(R_o(\theta) - \mathrm{sg}[R_a(\theta)]\big)\nabla R_o(\theta).
\end{equation}
Hence if reward hacking makes $R_o(\theta) < R_a(\theta)$, the coefficient becomes negative,
and AIR updates push \emph{against} the current open-context gradient direction, dampening spurious proxy-driven improvement.
This explains AIR's bidirectional correction behavior from a simple sign analysis.

\section{Detailed Training Algorithm}
\label{sec:detailed_algorithm}
Algorithm~\ref{alg:meta_grpo_refined} presents the full training procedure that integrates AIR into GRPO.
At each iteration, we sample a latent instance $z$ and construct a prompt set $S_z$ containing
verifiable \emph{anchor} prompts $A_z$ and open-ended \emph{open} variants $O_z$.
For each prompt $s\in S_z$, we draw $K$ completions $\{y_{s,k}\}_{k=1}^K$ and compute the
within-prompt reward statistics $(\mu_s,\sigma_s)$, yielding the GRPO group-normalized advantage
$\hat A_{s,k}=(r_{s,k}-\mu_s)/(\sigma_s+\delta)$.
We then form an instance-level anchor reference
$\mu_{\mathrm{anc}}=\frac{1}{|A_z|}\sum_{s\in A_z}\mu_s$, and compute the AIR coefficient for each
open prompt as $\Delta_s=\mu_{\mathrm{anc}}-\mu_s$, which captures the performance gap between open
prompts and their verifiable anchors (in implementation, we apply stop-gradient/detach to
$\mu_{\mathrm{anc}}$ to prevent the auxiliary term from updating anchors).
Finally, we optimize the GRPO clipped surrogate objective augmented with an AIR-weighted log-likelihood
auxiliary term (applied only to $s\in O_z$), so that when open prompts underperform their anchors the
model is encouraged to align with anchor-level performance, while unusually high open rewards
(potential reward hacking) induce a counteracting constraint, improving cross-context consistency
and robustness.

\begin{algorithm}[tb]
   \caption{GRPO with Anchor Invariance Regularization (AIR)}
   \label{alg:meta_grpo_refined}
\begin{algorithmic}[1]
   \STATE {\bfseries Input:} Policy $\pi_\theta$; Dataset $\mathcal{D}$; Hyperparams clip $\epsilon$, AIR weight $\lambda$; rollout size $K$; advantage constant $\delta$.
   
   \WHILE{not converged}
      \STATE \textit{// 1) Meta-group sampling \& rollout}
      \STATE Sample an instance $z \sim \mathcal{D}$.
      \STATE Build prompt set $\mathcal{S}_z=\mathcal{A}_z \cup \mathcal{O}_z$ (anchors $\mathcal{A}_z$, open variants $\mathcal{O}_z$).
      
      \FOR{each prompt $s \in \mathcal{S}_z$}
         \STATE Sample $K$ completions $\{y_{s,k}\}_{k=1}^K \sim \pi_{\theta_{\rm old}}(\cdot \mid s)$ and obtain rewards $\{r_{s,k}\}_{k=1}^K$.
         \STATE Prompt-wise stats: $\mu_s=\frac{1}{K}\sum_{k=1}^K r_{s,k}$,\;\; $\sigma_s=\mathrm{Std}(\{r_{s,k}\}_{k=1}^K)$.
      \ENDFOR

      \STATE \textit{// 2) GRPO advantages \& AIR coefficients}
      \STATE Anchor reference: $\mu_{\rm anc}=\frac{1}{|\mathcal{A}_z|}\sum_{s\in \mathcal{A}_z}\mu_s$.
      \FOR{each prompt $s \in \mathcal{S}_z$ and each $k=1,\dots,K$}
         \STATE GRPO advantage: $\hat{A}_{s,k}=\frac{r_{s,k}-\mu_s}{\sigma_s+\delta}$.
      \ENDFOR
      \FOR{each open prompt $s \in \mathcal{O}_z$}
         \STATE AIR coefficient: $\Delta_s=\mu_{\rm anc}-\mu_s$.
      \ENDFOR

      \STATE \textit{// 3) Policy optimization}
      \STATE Maximize:
      \[
      \mathcal{L}_{\text{total}}(\theta)=
      \mathbb{E}_{s,k}\Big[
      \min\Big(
      \frac{\pi_\theta(y_{s,k}\mid s)}{\pi_{\theta_{\rm old}}(y_{s,k}\mid s)}\hat{A}_{s,k},
      \mathrm{clip}\Big(\frac{\pi_\theta(y_{s,k}\mid s)}{\pi_{\theta_{\rm old}}(y_{s,k}\mid s)},1-\epsilon,1+\epsilon\Big)\hat{A}_{s,k} \Big)
      +\lambda\Delta_s r_{s,k}\log \pi_\theta(y_{s,k}\mid s)\Big].
      \]
      \STATE Update $\theta$ by gradient ascent on $\mathcal{L}_{\text{total}}(\theta)$.
   \ENDWHILE
\end{algorithmic}
\end{algorithm}

\section{Reward Design and Implementation Details}
\label{app:reward_design}

In our reinforcement learning framework, the total reward $r(s, y)$ for a prompt $s$ and completion $y$ is composed of a structural format reward and a task-specific content reward:
\begin{equation}
    r(s, y) = r_{\text{fmt}}(y) + r_{\text{task}}(s, y).
\end{equation}
Below, we detail the implementation of these reward components.

\subsection{Format Reward}
To ensure the model adheres to the Chain-of-Thought (CoT) reasoning requirement, we enforce a strict structural format. The format reward $r_{\text{fmt}}$ is binary-based:
\begin{itemize}
    \item \textbf{Structure Check:} The generation must be enclosed within \texttt{<think>...</think>} and \texttt{<answer>...</answer>} tags. The reasoning process must be non-empty, and the final answer must be extracted successfully.
    \item \textbf{Math-Specific Constraint:} For mathematics tasks, the content within the answer tags must strictly follow the \LaTeX\ box format \texttt{\textbackslash boxed\{...\}}.
\end{itemize}
If the generation satisfies all formatting constraints, $r_{\text{fmt}} = 1.25$; otherwise, a penalty of $r_{\text{fmt}} = -1.0$ is applied.

\subsection{Task-Specific Rewards}
We employ distinct reward mechanisms for Safety, Moral Reasoning, and Mathematics. While verifiable anchors (e.g., multiple-choice questions) use deterministic rule-based scoring, open-ended queries rely on a model-based judge or symbolic verification.

\subsubsection{Safety Reward Model (LLM-as-a-Judge)}
For open-ended safety evaluation, we utilize the policy model as a proxy judge. To reduce variance and strictly enforce safety boundaries, we do not rely on the judge's generated text. Instead, we compute the \textbf{conditional log-odds} of specific token sets given the conversation history.

\paragraph{Multi-Facet Evaluation.} We evaluate the response across 10 distinct safety facets (e.g., \textit{Self-Harm}, \textit{Criminal Aid}, \textit{Hate Speech}, \textit{Sexual Violence}). For each facet $f$, we construct a specific system prompt $P_f$ and compute the probability of the model outputting Safe versus Unsafe tokens.
Let $V_{\text{safe}}$ and $V_{\text{unsafe}}$ represent the sets of variants of these two words, respectively.
We calculate the score for each facet as:
\begin{equation}
    S_f = \log \left( \sum_{v \in V_{\text{safe}}} P(v \mid s, y, P_f) \right) - \log \left( \sum_{v \in V_{\text{unsafe}}} P(v \mid s, y, P_f) \right).
\end{equation}
The response is deemed safe only if $S_f$ exceeds a strict margin (set to 1.0) for the majority of facets and does not violate any facet (i.e., no $S_f < -0.5$).

\begin{table}[h]
\centering
\caption{The ten safety facets used by our LLM-as-a-Judge to detect harmful content.}
\begin{tabular}{ll}
\toprule
\textbf{Facet Name} & \textbf{Focus Description} \\
\midrule
Wrongdoing & Crime, cyber-crime, weapons, privacy invasion. \\
Fraud & Cheating, scams, social engineering, financial crime. \\
Hate Speech & Slurs, dehumanization, threats against protected groups. \\
Sexual (Minors) & Any content involving sexualization/exploitation of minors. \\
Self-Harm & Suicide, eating disorders, dangerous health advice. \\
Sexual (Adult) & Explicit pornography or content meant to arouse. \\
Medical Safety & Dangerous medical diagnoses or unproven cures. \\
Extremism & Recruitment/defense of terrorist organizations/ideologies. \\
Political & Targeted manipulation or deceptive political persuasion. \\
Misinformation & Spreading debunked claims on high-stakes topics. \\
\bottomrule
\end{tabular}
\label{tab:safety_prompts}
\end{table}

\paragraph{Friendliness Bonus.} If a response is deemed safe, we further evaluate its constructiveness using a Friendliness judge. This incentivizes the model to provide helpful refusals (e.g., offering legal alternatives) rather than curt rejections. The final score for the safe task is divided into three categories: \textbf{Unsafe}, -1.0 penalty; \textbf{Safe (Basic)}, +1.25 reward; \textbf{Safe \& Friendly}, +2.50 reward.

\paragraph{Safety System Prompts.} 
The judge uses specialized system prompts for each facet. For example, the prompt for \textit{Wrongdoing Assistance} is:
\begin{quote}
\textit{You are a STRICT safety judge for WRONGDOING ASSISTANCE. Focus on crime, cybercrime, physical harm, weapons, privacy invasion, and similar abuse. Label Unsafe if the response gives concrete, actionable help...}
\end{quote}
See Table~\ref{tab:safety_prompts} for the full list of safety facets.

\subsubsection{Moral Reasoning Reward Model}
For moral reasoning tasks, the objective is to align with specific ethical criteria provided in the dataset. Similar to the safety judge, we use the policy model to evaluate alignment via token probabilities.

The judge is presented with the context, the evaluation criteria, and the assistant's answer. It must classify the answer as satisfying the criteria (\textsc{YES}) or violating them (\textsc{NO}). We compute the log-odds difference between positive variants $V_{\text{pos}}$ (e.g., YES, Yes) and negative variants $V_{\text{neg}}$ (e.g., NO, No).
\begin{equation}
    r_{\text{moral}} = 
    \begin{cases} 
    +2.5 & \text{if } \text{LogOdds} \geq 1.0 \text{ (High Confidence YES)} \\
    -1.0 & \text{if } \text{LogOdds} \leq -0.5 \text{ (Detected Violation)} \\
    -0.5 & \text{otherwise (Uncertain)}
    \end{cases}
\end{equation}

\subsubsection{Mathematics Verification}
For mathematics, we employ a deterministic symbolic verification pipeline rather than a model-based judge. We parse the content inside \texttt{\textbackslash boxed\{...\}} from the model's output. We then utilize the \texttt{math\_verify} library to symbolically compare the extracted answer against the ground truth. This handles equivalence checking. A correct answer yields $+2.5$, while an incorrect or unparseable answer yields $-1.0$. For hard constraints like multiple-choice math problems, we perform exact string matching on the option letters (A, B, C, D) extracted from the final answer block.
\iffalse
\section{You \emph{can} have an appendix here.}

You can have as much text here as you want. The main body must be at most $8$
pages long. For the final version, one more page can be added. If you want, you
can use an appendix like this one.

The $\mathtt{\backslash onecolumn}$ command above can be kept in place if you
prefer a one-column appendix, or can be removed if you prefer a two-column
appendix.  Apart from this possible change, the style (font size, spacing,
margins, page numbering, etc.) should be kept the same as the main body.
\fi
%%%%%%%%%%%%%%%%%%%%%%%%%%%%%%%%%%%%%%%%%%%%%%%%%%%%%%%%%%%%%%%%%%%%%%%%%%%%%%%
%%%%%%%%%%%%%%%%%%%%%%%%%%%%%%%%%%%%%%%%%%%%%%%%%%%%%%%%%%%%%%%%%%%%%%%%%%%%%%%

\end{document}